\documentclass[runningheads]{llncs}
\usepackage{graphicx}

\usepackage{tikz}
\usepackage{comment}
\usepackage{amsmath,amssymb} 
\usepackage{color}
\usepackage{romannum}
\usepackage{dsfont}
\usepackage{multirow}
\usepackage{makecell}
\usepackage{hyperref}
\usepackage{orcidlink}

\usepackage[accsupp]{axessibility}  

\begin{document}
\pagestyle{headings}
\mainmatter
\def\ECCVSubNumber{5436}  

\title{Union-set Multi-source Model Adaptation for Semantic Segmentation\thanks{This study was partly supported by JSPS KAKENHI Grant Number JP21H03456 and conducted on the Data Science Computing System of Education and Research Center for Mathematical and Data Science, Hokkaido University.}} 

\titlerunning{Union-set Multi-source Model Adaptation for Semantic Segmentation}

\author{Zongyao Li\orcidlink{0000-0002-3300-1806} \and Ren Togo\orcidlink{0000-0002-4474-3995} \and Takahiro Ogawa\orcidlink{0000-0001-5332-8112} \and Miki Haseyama\orcidlink{0000-0003-1496-1761}}

\institute{Hokkaido University, Japan\\
\email{\{li,togo,ogawa,mhaseyama\}@lmd.ist.hokudai.ac.jp}}


\maketitle
\begin{abstract}
This paper solves a generalized version of the problem of multi-source model adaptation for semantic segmentation. Model adaptation is proposed as a new domain adaptation problem which requires access to a pre-trained model instead of data for the source domain. A general multi-source setting of model adaptation assumes strictly that each source domain shares a common label space with the target domain. As a relaxation, we allow the label space of each source domain to be a subset of that of the target domain and require the union of the source-domain label spaces to be equal to the target-domain label space. For the new setting named union-set multi-source model adaptation, we propose a method with a novel learning strategy named model-invariant feature learning, which takes full advantage of the diverse characteristics of the source-domain models, thereby improving the generalization in the target domain. We conduct extensive experiments in various adaptation settings to show the superiority of our method. The code is available at \url{https://github.com/lzy7976/union-set-model-adaptation}.
\keywords{Model adaptation, domain adaptation, semantic segmentation}
\end{abstract}

\section{Introduction}
Learning with unlabeled data is a long-term problem in the field of machine learning, and it has shown great significance more than ever since the rise of deep learning which heavily relies on well-labeled data. Due to the difficulty of unsupervised learning, some studies import labeled data from a different domain, and the problem accordingly becomes unsupervised domain adaptation (UDA)~\cite{ganin2015unsupervised,long2015learning}. Typically, a method of UDA borrows the knowledge from a labeled source domain for learning with an unlabeled target domain by using data of both the domains and reducing the domain gap~\cite{hoffman2018cycada}. However, considering the fact that regulations are increasingly and strictly constituted for protecting private data, sharing the source-domain data may be impractical in some applications. For such situations, an alternative way for borrowing the source-domain knowledge without direct use of the source-domain data is necessary.
\par
Model adaptation~\cite{li2020model} (also named source-free domain adaptation) has been proposed for solving the above problem. As a derivative of UDA, model adaptation replaces the source-domain data with a source-domain pre-trained model and is thus no longer limited by the restriction of data sharing. Since private information can be hardly recovered from the pre-trained models, model adaptation faces less limitations and is practical in a wider range of applications than traditional UDA. Furthermore, in addition to the less difficulty of getting the access permission, pre-trained models also require considerably less storage size than training data, and therefore using multiple pre-trained models of different source domain, i.e., multi-source model adaptation (MSMA), seems to be a cost-efficient option when multiple appropriate source domains exist. However, unlike multi-source UDA which has been widely studied~\cite{zhao2018adversarial,zhao2019multi,he2021multi}, the study on MSMA is insufficient despite its promising prospect. To the best of our knowledge, only one work studied on MSMA for image classification~\cite{ahmed2021unsupervised}.
\par

This paper focuses on the problem of MSMA for semantic segmentation which still remains to be solved. Similar to multi-source UDA, in a general MSMA setting, a common label space is expected to be shared by all the source domains and the target domain, which is a too stringent assumption to be practical in some real-world scenarios. Therefore, in this paper, we relax the requirement for the source-domain label spaces and propose a generalized version of MSMA named union-set multi-source model adaptation (US-MSMA). Specifically, in our US-MSMA setting, the union of all the source-domain label spaces instead of the label space of each source domain is required to be equal to the target domain label space. In other words, the label space of each source domain is just expected to be a subset but not necessarily the same as that of the target domain. Such a relaxation considerably extends the applicability of MSMA. Moreover, it also allows selection of source domains from a larger candidate set which may improve the adaptation performance. For example, a high-performance model that is trained in a source domain with high-quality labels of part of the target domain classes, can be used in the training of US-MSMA and contributes to improving the adaptation performance for certain classes. The generalized multi-source setting is especially compatible with model adaptation due to the low cost for introducing pre-trained models.
\par
For handling the problem of US-MSMA for semantic segmentation, we propose a two-stage method which consists of a model adaptation stage and a model integration stage. For the model adaptation stage, we propose a novel learning strategy, model-invariant feature learning. Specifically, to take full advantage of diverse characteristics of the source-domain pre-trained models, we train the source-domain models to produce target-domain features with similar distributions which are referred to as model-invariant features. The conception of the model-invariant feature learning is to reduce the domain biases and improve the generalization ability of the source-domain models by harmonizing the model characteristics derived from different source domains. Moreover, to obtain predictions in the target-domain label space, we introduce a classifier ensemble strategy which combines predictions of all the classifiers of the source-domain models as the complete prediction. To integrate the adapted source-domain models, we further introduce the model integration stage which distills knowledge of the adapted models to train a final model. We validate the effectiveness of our method in various situations of the union-set multi-source setting by conducting extensive experiments.
\par
This paper's contributions are summarized as follows.
\begin{itemize}
\item We propose the problem setting of US-MSMA, which relaxes the requirement for the source-domain label spaces in the general multi-source setting and is thus applicable to a wider range of practical scenarios.
\item We propose a two-stage method to handle the problem of US-MSMA for semantic segmentation. In the first stage, we propose the novel model-invariant feature learning for better generalization in the target domain. And as the second stage, we introduce the model integration to train a final model which absorbs knowledge from the adapted source-domain models.
\item We conduct experiments in extensive adaptation settings that use several source-domain sets with different label space settings. Experimental results demonstrate the superiority of our method to previous adaptation methods.
\end{itemize}

\section{Related Works}
\subsection{UDA for Semantic Segmentation}
UDA for semantic segmentation has been widely studied in recent years. Typical technologies used in the methods mainly include image-to-image translation~\cite{hoffman2018cycada,li2019bidirectional,chen2019crdoco,wu2018dcan,yang2020fda}, adversarial learning~\cite{zhang2018fully,tsai2018learning,tsai2019domain,du2019ssf,luo2019taking}, and semi-supervised learning~\cite{zou2018unsupervised,zheng2021rectifying,choi2019self,vu2019advent,chen2019domain}. Image-to-image translation is used for reducing the visual domain gap by modifying some image characteristics and is mainly performed with a GAN~\cite{goodfellow2014generative}-based model~\cite{zhu2017unpaired}. Adversarial learning introduces a domain discriminator for recognizing the domain of intermediate features~\cite{zhang2018fully} or final outputs~\cite{tsai2018learning}. Training the segmentation network and the discriminator against each other can align the feature distributions of the domains. Semi-supervised learning technologies can be readily applied to UDA due to the similar problem setting, including pseudo-label learning~\cite{zou2018unsupervised,zheng2021rectifying}, self-ensembling~\cite{choi2019self}, and entropy minimization~\cite{vu2019advent,chen2019domain}. In this paper, we also use the pseudo-label learning as the baseline of our method and introduce the entropy minimization for further improving the adaptation performance.
\subsection{Multi-source UDA}
Multi-source UDA methods have been developed for image classification~\cite{hoffman2018algorithms,peng2019moment,zhao2018adversarial}, semantic segmentation~\cite{zhao2019multi,he2021multi}, and object detection~\cite{yao2021multi}. The technologies used in single-source UDA play a dominant role also in multi-source UDA. Zhao et al.~\cite{zhao2018adversarial} introduce multiple domain discriminators on the top of the feature extractor and conduct the adversarial learning between the feature extractor and the discriminators via a gradient reversal layer. Peng et al.~\cite{peng2019moment} align moments of feature distributions of the source domains and the target domain to transfer the source-domain knowledge. Zhao et al.~\cite{zhao2019multi} align the domains at both the pixel level and the feature level with the image-to-image translation and the adversarial learning. He et al.~\cite{he2021multi} also perform the pixel-level adaptation and further train with target-domain pseudo labels. Yao et al.~\cite{yao2021multi} train a domain-adaptive object detector with multiple source subnets and obtain a target subnet by weighting and combining the parameters of the source subnets. The above methods are all developed for the general multi-source setting which is a particular case of our union-set multi-source setting.
\subsection{Model Adaptation}
Previous studies on model adaptation mainly focus on image classification, using some technologies similar to those used in UDA, such as generative models~\cite{li2020model}, adversarial learning~\cite{xia2021adaptive}, information maximization~\cite{liang2020we}, and class prototypes~\cite{yang2020unsupervised}. As to semantic segmentation, Liu et al.~\cite{liu2021source} generate fake source-domain samples with real source-domain distribution to transfer the source-domain knowledge and introduce a patch-level self-supervision module for target-domain pseudo labels. Fleuret et al.~\cite{fleuret2021uncertainty} reduce uncertainty of predictions from multiple classifiers suffering from random noise to enhance robustness of the learned feature representation. Stan et al.~\cite{stan2021unsupervised} learn a prototypical distribution to encode source-domain knowledge and align the distributions across domains with the learned distribution. Model adaptation in the multi-source setting is studied poorly, with only one work~\cite{ahmed2021unsupervised} for image classification to the best our knowledge. Ahmed et al.~\cite{ahmed2021unsupervised} learn an optimal combination of multiple source-domain models with trainable weights, whereas the inference time is increased by several times due to the model combination.

\section{Proposed Method}
\subsection{Problem Setting of US-MSMA}
First, we detail the US-MSMA problem setting of our method. Let $\{D^S_i\}_{i=1}^k$ denote $k$ labeled source domains with class sets $\{\Phi ^S_i\}_{i=1}^k$ and $D^T$ the unlabeled target domain with a class set $\Phi ^T$. Different from the general multi-source setting in which each of $\{\Phi ^S_i\}_{i=1}^k$ must be absolutely the same as $\Phi ^T$, in our US-MSMA setting, the union of $\{\Phi ^S_i\}_{i=1}^k$ is assumed to be equal to $\Phi ^T$, i.e., $\Phi ^S_1\bigcup \Phi ^S_2 \dots \bigcup \Phi ^S_k=\Phi ^T$. Given access to unlabeled data $\{x^T_i\}_{i=1}^n$ of $D^T$ and $k$ models $\{M_i\}_{i=1}^k$ pre-trained with $\{D^S_i\}_{i=1}^k$ respectively, we aim to obtain a model that learns knowledge transferred from $\{D^S_i\}_{i=1}^k$ to $D^T$ and consequently achieves reasonable performance in $D^T$.

\begin{figure}[t]
\centering
\includegraphics[width=12.0cm]{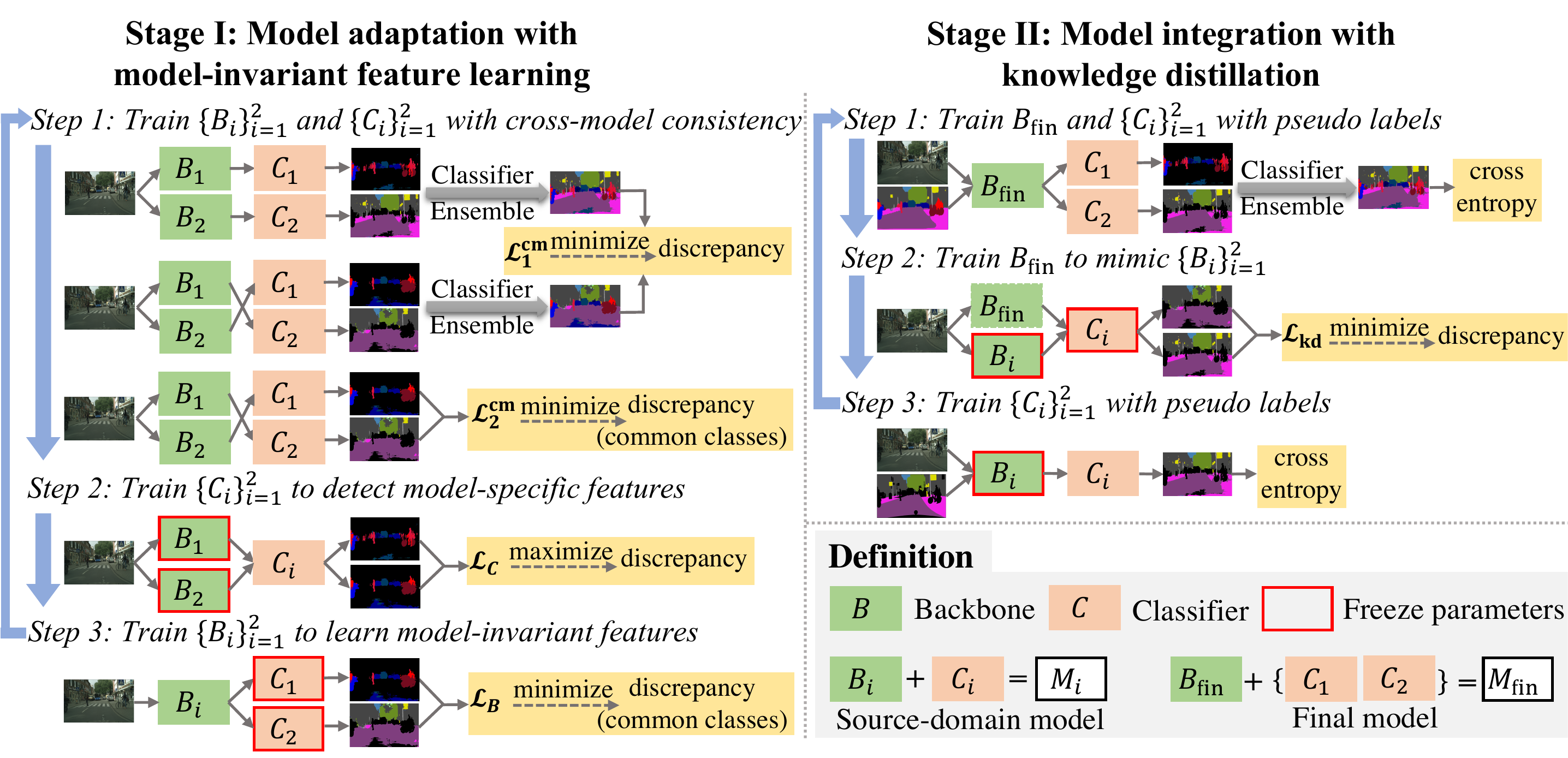}
\caption{An overview of the proposed method. For the ease of understanding, we show the case of using two source-domain models. The self-training with pseudo labels in Stage \Romannum{1} is omitted in the figure.}
\end{figure}

\subsection{Overview of The Proposed Method}
\subsubsection{Two-stage architecture}
Figure 1 shows an overview of the proposed method which consists of the following two stages: a model adaptation stage and a model integration stage. In Stage \Romannum{1}, we conduct the model adaptation by retraining the pre-trained source-domain models $\{M_i\}_{i=1}^k$ with the target domain $D^T$. The source-domain knowledge is transferred to the target domain by training $\{M_i\}_{i=1}^k$ with pseudo labels of $D^T$ in the manner of self-training (omitted in Fig. 1), and we improve the adaptation with the model-invariant feature learning. Then, in Stage \Romannum{2}, we train a final model $M_{\rm fin}$ by distilling and integrating knowledge from $\{M_i\}_{i=1}^k$ trained in Stage \Romannum{1}. As defined in Fig. 1, $\{M_i\}_{i=1}^k$ are individual models composed of a backbone $B_i$ and a classifier $C_i$, and $M_{\rm fin}$ is an ensemble model composed of an integration backbone $B_{\rm fin}$ and all the classifiers $\{C_i\}_{i=1}^k$. The classifiers of the ensemble model are combined with a classifier ensemble strategy described below.
\subsubsection{Classifier ensemble}
In both Stages \Romannum{1} and \Romannum{2}, we predict the probability distribution over all the target-domain classes $\Phi ^T$ with the classifiers $\{C_i\}_{i=1}^k$. However, since the source-domain class sets $\{\Phi ^S_i\}_{i=1}^k$ are not necessarily equal to $\Phi ^T$, the prediction over $\Phi ^T$ may not be available from any individual $C_i$. And due to the possibly different class sets, the predictions of $\{C_i\}_{i=1}^k$ cannot be simply averaged. Therefore, to obtain the complete prediction, we use a classifier ensemble strategy which simultaneously combines and averages the predictions of $\{C_i\}_{i=1}^k$. Specifically, we calculate the unnormalized logits of the complete prediction by averaging the output logits of each class over $\{C_i\}_{i=1}^k$ as the following equation:
\begin{equation}
l_c(\cdot)=\frac{1}{\sum_{i=1}^k\mathds{1}(c\in \Phi _i^S)}\sum_{i=1}^kC_{i,c}(\cdot),\ \forall c\in \Phi ^T,
\end{equation}
where $\mathds{1}(\cdot)$ denotes the indicator function, and $C_{i,c}(\cdot)$ denotes the logits of class $c$ predicted by $C_i$ if $c\in \Phi _i^S$ and is regarded as zero otherwise. The calculated logits are then normalized with the Softmax function as the predicted probability distribution in the target-domain label space. In such a manner, we obtain the prediction in the target-domain label space, using the classifiers $\{C_i\}_{i=1}^k$ with incomplete class sets instead of training a new classifier. The classifier ensemble operation is hereinafter denoted by ${\rm En}(\cdot)$.

\subsection{Stage \Romannum{1}: Model Adaptation with Model-invariant Feature Learning}
As mentioned above, we conduct the model adaptation in Stage \Romannum{1} on the basis of the self-training with the pseudo labels and introduce the model-invariant feature learning to improve the adaptation performance. The model-invariant feature learning aims to reduce the domain biases of $\{M_i\}_{i=1}^k$. Since the pre-trained $\{M_i\}_{i=1}^k$ initially have diverse characteristics derived from the source domains, the domain bias of each $M_i$ can be reduced by harmonizing the characteristics of $\{M_i\}_{i=1}^k$. We realize it by training the backbones $\{B_i\}_{i=1}^k$ to produce features with similar distributions which we refer to as model-invariant features. The model-invariant features are learned by a iterative process composed of three steps: the first step for cross-model consistency and the subsequent two steps for adversarial learning between the backbones $\{B_i\}_{i=1}^k$ and the classifiers $\{C_i\}_{i=1}^k$, as shown in the left side of Fig. 1. We detail each component of Stage \Romannum{1} as follows.
\subsubsection{Self-training with pseudo labels}
To transfer the source-domain knowledge to the target domain, we perform the self-training for each of the individual models $\{M_i\}_{i=1}^k$ and also ensemble models that are composed of a backbone $B_i$ ($i=1,\dots ,k$) and all the classifiers $\{C_i\}_{i=1}^k$. To this end, we use the pre-trained $\{M_i\}_{i=1}^k$ to generate pseudo labels $\{y_i\}_{i=1}^k$ in the source-domain label spaces and pseudo labels $y^T$ in the target-domain label space for each image of $D^T$. To generate the pseudo labels in the target-domain label space, we combine and average the predictions of $\{M_i\}_{i=1}^k$ but not like the classifier ensemble described above since $\{M_i\}_{i=1}^k$ are pre-trained independently. Specifically, we first cast the probability distributions predicted by $\{M_i\}_{i=1}^k$ over the target-domain label space as $p_{i,c}(\cdot)=M_{i,c}(\cdot)$ if $c\in \Phi ^S_i$ and $p_{i,c}(\cdot)=\frac{M_{i,0}(\cdot)}{\sum_{c'\in \Phi ^T}\mathds{1}(c'\notin \Phi ^S_i)}$ otherwise, where $M_{i,c}(\cdot)$ is the probability of class $c$ predicted by $M_i$ and $M_{i,0}(\cdot)$ is the probability of the other classes not in $\Phi ^S_i$. Then we average the probability distributions $\{p_i\}_{i=1}^k$ and assign pseudo labels according to the average prediction. In the self-training, we use the cross-entropy loss function ${\rm CE}(logits,target)$ to train $\{M_i\}_{i=1}^k$ as follows:
\begin{equation}
\mathcal{L}_{\rm pl}=\mathbb{E}_{x^T\in D^T}\sum_{i=1}^k[{\rm CE}(C_i(B_i(x^T)),y_i)+{\rm CE}({\rm En}(\{C_j(B_i(x^T))\}_{j=1}^k),y_T)],
\end{equation}
where ${\rm En}(\cdot)$ is the classifier ensemble operation described in Section 3.2.
\subsubsection{Cross-model consistency}
On the basis of the assumption that a backbone that produces model-invariant features should be compatible with any classifier, we randomly recombine $\{B_i\}_{i=1}^k$ and $\{C_i\}_{i=1}^k$ as $k$ new models $\{C_{{\rm ma}(i)}(B_i(\cdot))\}_{i=1}^k$ where ${\rm ma}(i)$ is the index of the classifier matched with $B_i$. The recombined models are trained in terms of cross-model consistency which includes overall consistency of the ensemble predictions and per-class consistency of the logits output by individual classifiers. For the overall consistency, we perform the classifier ensemble operation for predictions of the original models $\{C_i(B_i(\cdot))\}_{i=1}^k$ and the recombined models $\{C_{{\rm ma}(i)}(B_i(\cdot))\}_{i=1}^k$ and minimize the discrepancy between the ensemble predictions with the following loss function:
\begin{equation}
\mathcal{L}^{\rm cm}_1=\mathbb{E}_{x^T\in D^T}||\sigma({\rm En}(\{C_i(B_i(x^T))\}_{i=1}^k))-\sigma({\rm En}(\{C_{{\rm ma}(i)}(B_i(x^T))\}_{i=1}^k))||_1,
\end{equation}
where $\sigma(\cdot)$ is the Softmax function. For the per-class consistency, we train the recombined models to output consistent logits of each class. We calculate the average logits of each class and minimize the discrepancy between the output logits and the average logits as follows:
\begin{equation}
\mathcal{L}^{\rm cm}_2=\mathbb{E}_{x^T\in D^T}\sum_{i=1}^k\sum_c^{\Phi ^S_i}||C_{{\rm ma}(i),c}(B_i(x^T))-\delta_c(x^T)||_1,
\end{equation}
where $\delta_c(\cdot)$ is the average logits of class $c$ and calculated with the following equation:
\begin{equation}
\delta_c(\cdot)=\frac{1}{\sum_{i=1}^k\mathds{1}(c\in \Phi^S_i)}\sum_{i=1}^kC_{{\rm ma}(i),c}(B_i(\cdot)),
\end{equation}
where $C_{{\rm ma}(i),c}(\cdot)$ is the logits of class $c$ output by $C_{{\rm ma}(i)}$ if $c\in \Phi^S_{{\rm ma}(i)}$ and zero otherwise. By recombining and training the models with $\mathcal{L}^{\rm cm}_1$ and $\mathcal{L}^{\rm cm}_2$ for the overall and the per-class consistency respectively, the features produced by the backbones are constrained to have similar distributions, i.e., to be model-invariant.
\subsubsection{Adversarial learning}
In addition to the cross-model consistency, we further introduce adversarial learning between the backbones $\{B_i\}_{i=1}^k$ and the classifiers $\{C_i\}_{i=1}^k$ to enhance the model-invariant feature learning. Specifically, the adversarial learning consists of two steps: training $\{C_i\}_{i=1}^k$ to detect model-specific features and training $\{B_i\}_{i=1}^k$ to produce model-invariant features. The two steps are performed iteratively, and the parameters of $\{B_i\}_{i=1}^k$ ($\{C_i\}_{i=1}^k$) are frozen while updating those of $\{C_i\}_{i=1}^k$ ($\{B_i\}_{i=1}^k$).
\par
For the training of $C_i$ ($i=1,\dots ,k$), we feed the features from $B_i$ and $B_j$ ($j\neq i$) into $C_i$ and maximize the discrepancy between the predictions by minimizing the following loss function:
\begin{equation}
\mathcal{L}_C=\mathbb{E}_{x^T\in D^T}\sum_{i=1}^k[\sum_{j=1}^k-||C_i(B_i(x^T))-C_i(B_j(x^T))||_1+{\rm CE}(C_i(B_i(x^T)),y_i)],
\end{equation}
where we add a cross-entropy term to prevent the recognition ability of $\{C_i\}_{i=1}^k$ from degradation while maximizing the discrepancy. The features from $B_j$ that induce inconsistent predictions with those using the features from $B_i$, are considered domain-specific and detected by updating $C_i$ with $\mathcal{L}_C$.
\par
We train $\{B_i\}_{i=1}^k$ with a loss that calculates the discrepancy between the output logits and the average logits with features from each of the backbones respectively, as defined in the following equation:
\begin{equation}
\mathcal{L}_B=\mathbb{E}_{x^T\in D^T}\sum_{i=1}^k\sum_{j=1}^k[\sum_c^{\Phi ^S_j}||C_{j,c}(B_i(x^T))-\delta_{i,c}(x^T)||_1+{\rm CE}(C_j(B_i(x^T)),y_j)],
\end{equation}
where $\delta_{i,c}(\cdot)$ is the average logits of class $c$ using the features from $B_i$ and calculated as follows:
\begin{equation}
\delta_{i,c}(\cdot)=\frac{1}{\sum_{j=1}^k\mathds{1}(c\in \Phi^S_j)}\sum_{j=1}^kC_{j,c}(B_i(\cdot)).
\end{equation}
By minimizing the L1-norm term of $\mathcal{L}_B$, each of the backbones is trained to produce features that induce per-class consistent logits from different classifiers and are thus considered domain-invariant. However, since the L1-norm term involves only the classes shared by multiple classifiers, we additionally add a cross-entropy term to train each backbone to be compatible with all the classifiers.
\par
Unlike the typical adversarial learning, we conduct the adversarial learning between two groups of the backbones $\{B_i\}_{i=1}^k$ and the classifiers $\{C_i\}_{i=1}^k$ rather than two specific opponents. Moreover, we train $\{B_i\}_{i=1}^k$ with a loss dissimilar to that for $\{C_i\}_{i=1}^k$, instead of minimizing the term $||C_i(B_i(x^T))-C_i(B_j(x^T))||_1$ of $\mathcal{L}_C$ which may lead to excessive similarity among $\{B_i\}_{i=1}^k$. The adversarial learning is compatible with the cross-model consistency and can further enhance the model-invariance of the learned features.
\par
With the components described above, the training of Stage \Romannum{1} is performed by repeating a process of three steps. Step 1 simultaneously updates $\{B_i\}_{i=1}^k$ and $\{C_i\}_{i=1}^k$ with the loss $\mathcal{L}_{\rm pl}$ of the self-training and the losses $\mathcal{L}^{\rm cm}_1$ and $\mathcal{L}^{\rm cm}_2$ of the cross-model consistency. Then Step 2 and Step 3 updates $\{C_i\}_{i=1}^k$ and $\{B_i\}_{i=1}^k$ with $\mathcal{L}_C$ and $\mathcal{L}_B$, respectively.

\subsection{Stage \Romannum{2}: Model Integration with Knowledge Distillation}
The models adapted in Stage \Romannum{1} have slightly different performances, and it is generally unknown which one would perform best in the test. To achieve the best performance not inferring with all the models, we introduce the model integration stage that distills the knowledge of $\{M_i\}_{i=1}^k$ and trains a final model $M_{\rm fin}$. $M_{\rm fin}$ is composed of an integration backbone $B_{\rm fin}$ and all the classifiers $\{C_i\}_{i=1}^k$, and the ensemble prediction ${\rm En}(\{C_i(B_{\rm fin}(\cdot))\}_{i=1}^k)$ is regarded as the final prediction. The parameters of the backbones $\{B_i\}_{i=1}^k$ are frozen in this stage.
\par
Similar to Stage \Romannum{1}, $M_{\rm fin}$ is trained with a loss that updates $B_{\rm fin}$ and $\{C_i\}_{i=1}^k$ together and losses that update $B_{\rm fin}$ and $\{C_i\}_{i=1}^k$ respectively. Specifically, $B_{\rm fin}$ and $\{C_i\}_{i=1}^k$ are trained together by minimizing ${\rm CE}({\rm En}(\{C_i(B_{\rm fin}(\cdot))\}_{i=1}^k),y_T)$, the cross-entropy loss for the ensemble prediction using pseudo labels that are generated with $\{M_i\}_{i=1}^k$ trained in Stage \Romannum{1}. Then, $B_{\rm fin}$ is trained to mimic $\{B_i\}_{i=1}^k$ with a knowledge distillation loss defined as follows:
\begin{equation}
\mathcal{L}_{\rm kd}=\mathbb{E}_{x^T\in D^T}\sum_{i=1}^k{\rm KLD}(C_i(B_{\rm fin}(x^T)),C_i(B_i(x^T))),
\end{equation}
where ${\rm KLD}(input, target)$ is the Kullback–Leibler divergence function. Moreover, to maintain the compatibility of $\{B_i,C_i\}_{i=1}^k$, we train $\{C_i\}_{i=1}^k$ by minimizing $\sum_{i=1}^k{\rm CE}(C_i(B_i(x^T)),y_i)$ for the predictions of individual classifiers. By training with the above losses, $M_{\rm fin}$ absorbs the knowledge of $\{M_i\}_{i=1}^k$ trained in Stage \Romannum{1} and consequently outperforms all of $\{M_i\}_{i=1}^k$. Moreover, since $\{C_i\}_{i=1}^k$ are very light networks, the inference speed of $M_{\rm fin}$ is almost the same as that of a single model of $\{M_i\}_{i=1}^k$.

\section{Experiments}
\subsection{Implementation Details}
As the segmentation network in all the experiments, we used Deeplab V2~\cite{chen2017deeplab} with ResNet101~\cite{he2016deep} of which the ResNet101 backbone and the atrous spatial pyramid pooling (ASPP) classifier were used as the backbones $\{B_i\}_{i=1}^k$ and the classifiers $\{C_i\}_{i=1}^k$, respectively. We trained the networks with a stochastic gradient descent (SGD) optimizer with an initial learning rate of 2.5$\times$10$^{-4}$ for 80,000 iterations. During the training, the learning rate was decreased with the poly policy with a power of 0.9, and the mini-batch size was set to 1. We used an equal weight of 1.0 for all the losses. Moreover, we also introduced a maximum squares loss~\cite{chen2019domain} which minimizes the uncertainty of predictions, as a supplement to the losses $\mathcal{L}_{\rm pl}$ and $\mathcal{L}_B$ in Stage \Romannum{1} and also the loss ${\rm CE}({\rm En}(\{C_i(B_{\rm fin}(\cdot))\}_{i=1}^k),y_T)$ in Stage \Romannum{2} to improve the adaptation performance.

\begin{table}[t]
\centering
\caption{Class distributions of the non-overlapping setting and the partly-overlapping setting. The target domain contains 19 classes: road, sidewalk, building, wall, fence, pole, traffic light, traffic sign, vegetation, terrain, sky, person, rider, car, truck, bus, train, motorcycle, bicycle.}
\setlength{\tabcolsep}{1.3pt}
\begin{tabular}{|c|c|c|ccccccccccc|cccccccc|}
\hline
\multicolumn{2}{|c|}{} & & \multicolumn{11}{c|}{Background classes} & \multicolumn{8}{c|}{Foreground classes} \\
\multicolumn{2}{|c|}{Setting} & \makecell[c]{Source\\domain} & \rotatebox{90}{road} & \rotatebox{90}{side.} & \rotatebox{90}{buil.} & \rotatebox{90}{wall} & \rotatebox{90}{fence} & \rotatebox{90}{pole} & \rotatebox{90}{t-lig.} & \rotatebox{90}{t-sign} & \rotatebox{90}{vege.} & \rotatebox{90}{terr.} & \rotatebox{90}{sky} & \rotatebox{90}{pers.} & \rotatebox{90}{rider} & \rotatebox{90}{car} & \rotatebox{90}{truck} & \rotatebox{90}{bus} & \rotatebox{90}{train} & \rotatebox{90}{moto.} & \rotatebox{90}{bicy.} \\
\hline
\multirow{7}*{\makecell[c]{Non-\\overlapping}} & \multirow{2}*{$S$+$G$} & $S$ & & & & & & & & & & & & $\checkmark$ & $\checkmark$ & $\checkmark$ & $\checkmark$ & $\checkmark$ & $\checkmark$ & $\checkmark$ & $\checkmark$ \\
& & $G$ & $\checkmark$ & $\checkmark$ & $\checkmark$ & $\checkmark$ & $\checkmark$ & $\checkmark$ & $\checkmark$ & $\checkmark$ & $\checkmark$ & $\checkmark$ & $\checkmark$ & & & & & & & & \\
\cline{2-22}
& \multirow{2}*{$S(G)$+$T$} & $S(G)$ & & & & & & & & & & $\checkmark$ & & $\checkmark$ & $\checkmark$ & $\checkmark$ & $\checkmark$ & $\checkmark$ & $\checkmark$ & $\checkmark$ & $\checkmark$ \\
& & $T$ & $\checkmark$ & $\checkmark$ & $\checkmark$ & $\checkmark$ & $\checkmark$ & $\checkmark$ & $\checkmark$ & $\checkmark$ & $\checkmark$ & & $\checkmark$ & & & & & & & & \\
\cline{2-22}
& \multirow{3}*{$S$+$G$+$T$} & $S$ & & & & & & & & & & & & & & $\checkmark$ & $\checkmark$ & $\checkmark$ & $\checkmark$ & & \\
& & $G$ & $\checkmark$ & $\checkmark$ & $\checkmark$ & $\checkmark$ & $\checkmark$ & $\checkmark$ & $\checkmark$ & $\checkmark$ & $\checkmark$ & $\checkmark$ & $\checkmark$ & & & & & & & & \\
& & $T$ & & & & & & & & & & & & $\checkmark$ & $\checkmark$ & & & & & $\checkmark$ & $\checkmark$ \\
\hline
\multirow{4}*{\makecell[c]{Partly-\\overlapping}} & \multirow{2}*{$S$+$G$} & $S$ & $\checkmark$ & $\checkmark$ & $\checkmark$ & $\checkmark$ & $\checkmark$ & $\checkmark$ & $\checkmark$ & $\checkmark$ & $\checkmark$ & $\checkmark$ & $\checkmark$ & & & $\checkmark$ & $\checkmark$ & $\checkmark$ & $\checkmark$ & & \\
& & $G$ & $\checkmark$ & $\checkmark$ & $\checkmark$ & $\checkmark$ & $\checkmark$ & $\checkmark$ & $\checkmark$ & $\checkmark$ & $\checkmark$ & $\checkmark$ & $\checkmark$ & $\checkmark$ & $\checkmark$ & & & & & $\checkmark$ & $\checkmark$ \\
\cline{2-22}
& \multirow{2}*{$S(G)$+$T$} & $S(G)$  & $\checkmark$ & $\checkmark$ & $\checkmark$ & $\checkmark$ & $\checkmark$ & $\checkmark$ & $\checkmark$ & $\checkmark$ & $\checkmark$ & $\checkmark$ & $\checkmark$ & & & $\checkmark$ & $\checkmark$ & $\checkmark$ & $\checkmark$ & & \\
& & $T$ & $\checkmark$ & $\checkmark$ & $\checkmark$ & $\checkmark$ & $\checkmark$ & $\checkmark$ & $\checkmark$ & $\checkmark$ & $\checkmark$ & & $\checkmark$ & $\checkmark$ & $\checkmark$ & & & & & $\checkmark$ & $\checkmark$ \\
\hline
\end{tabular}
\end{table}

\subsection{Datasets and Adaptation Settings}
We used Synscapes dataset~\cite{wrenninge2018synscapes}, GTA5 dataset~\cite{richter2016playing}, and Synthia dataset~\cite{ros2016synthia} as the source domains and Cityscapes dataset~\cite{cordts2016cityscapes} as the target domain. All the source domains are synthetic datasets that are composed of photo-realistic images of street scenes, and Cityscapes is a real-world dataset of street-scene images. Synscapes, GTA5, and Cityscapes share a common label space containing 19 classes, while Synthia shares a subset containing 16 classes with the others. Synscapes, GTA5, and Synthia are hereinafter referred to as $S$, $G$, and $T$, respectively.
\par
We conducted experiments with all the possible combinations of the source domains including $S$+$G$, $S$+$T$, $G$+$T$, and $S$+$G$+$T$. Furthermore, since our union-set multi-source setting is flexible in terms of the source-domain label spaces, we evaluated our method in various settings of the source-domain label spaces to demonstrate its effectiveness in a wide range of scenarios. Specifically, we conducted experiments in three label space settings: non-overlapping, partly-overlapping, and fully-overlapping. In the non-overlapping setting, the target-domain classes are divided into subsets with no common classes and assigned to the source domains. In the partly-overlapping setting, background classes are shared as common classes, while foreground classes are divided for the source domains. We detail the class distributions of the non-overlapping setting and the partly-overlapping setting in Table 1. In the fully-overlapping setting, all classes are shared expect three classes (terrain, truck, train) that are absent in Synthia, but we still involved these classes in the experiments of $S$+$T$ and $G$+$T$.

\begin{table}[t]
\centering
\caption{Results in the non-overlapping setting. ST: self-training. CMC: cross-model consistency. ADV: adversarial learning. MSL: maximum squares loss. MI: model integration. PM: proposed method.}
\setlength{\tabcolsep}{3.0pt}
\begin{tabular}{|l|cccc|}
\hline
Method & $S$+$G$ & $S$+$T$ & $G$+$T$ & $S$+$G$+$T$ \\
\hline
ST & 42.3 & 38.8 & 35.8 & 40.4 \\
ST+CMC & 43.2 & 39.1 & 36.1 & 40.6 \\
ST+CMC+ADV & 44.0 & 39.8 & 36.4 & 41.5 \\
ST+CMC+ADV+MSL (\textbf{=Stage \Romannum{1} of PM}) & 45.8 & 41.4 & 37.2 & 42.1 \\
ST+CMC+ADV+MSL+MI (\textbf{=PM}) & \textbf{46.6} & \textbf{42.3} & \textbf{37.9} & \textbf{44.2} \\
\hline
SSMA~\cite{fleuret2021uncertainty} & 43.5 & 40.6 & 37.0 & 41.9 \\
MSMA~\cite{ahmed2021unsupervised} & 30.2 & 26.0 & 20.7 & 22.7 \\
UDA~\cite{tsai2018learning} & 45.7 & 39.2 & 35.9 & 41.1 \\
\hline
\end{tabular}
\end{table}

\begin{figure}[t]
\centering
\includegraphics[width=12.0cm]{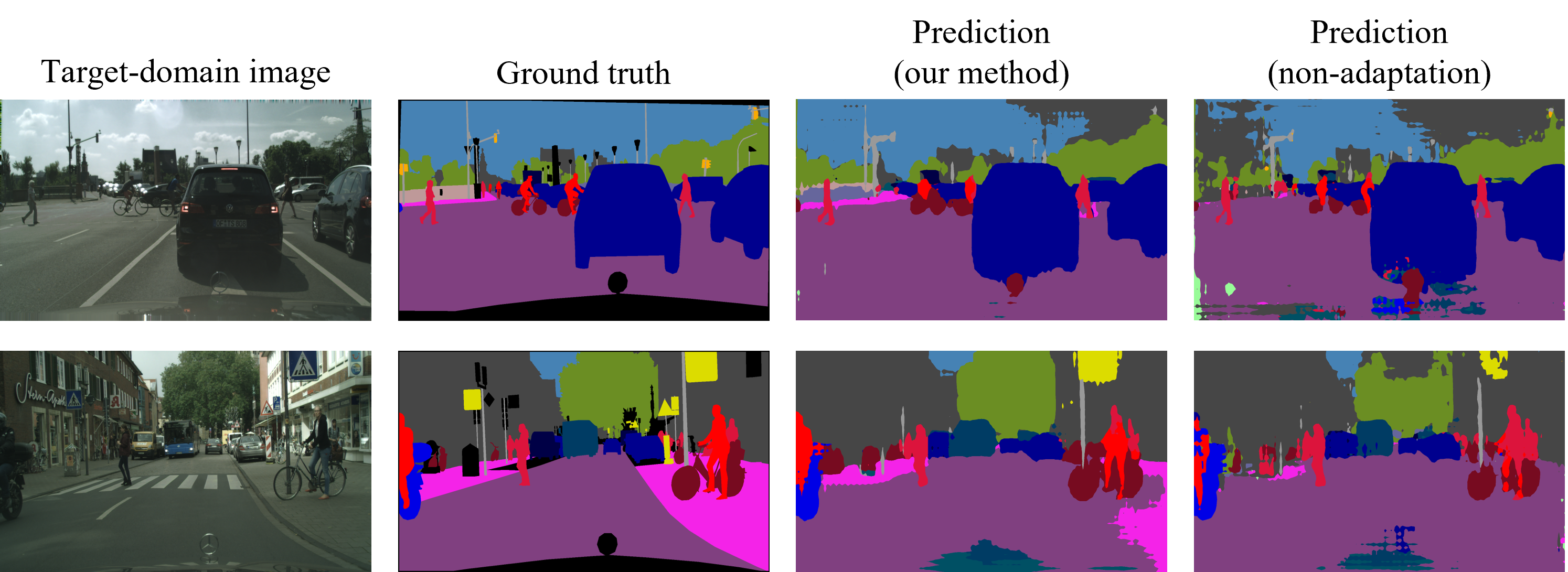}
\caption{Examples of the qualitative results of our method and the source-domain models without adaptation in the non-overlapping setting of $S$+$G$.}
\end{figure}

\subsection{Results in Non-overlapping Setting}
Since no existing methods are proposed for such a problem setting, we slightly modified three methods proposed for related problem settings for comparisons: a UDA method~\cite{tsai2018learning} for semantic segmentation, a single-source model adaptation (SSMA) method~\cite{fleuret2021uncertainty} for semantic segmentation, and an MSMA method~\cite{ahmed2021unsupervised} for image classification. For the UDA method based on adversarial learning, we used domain-specific discriminators and classifiers for the source domains and trained one shared backbone. For both the UDA method and the SSMA method, the complete predictions were obtained by first casting the predictions in the source-domain label spaces over the target-domain label space and then averaging the predictions, in the same manner as that for generating the pseudo labels. For the MSMA method which learns a set of weights for combining the models, we still cast the prediction of each model over the target-domain label space in the same manner and calculated the weighted average prediction with the learned weights. Moreover, for explicit comparisons with the SSMA method and the MSMA method, we performed the training of the methods using the same maximum squares loss and the cross-entropy loss with the same pseudo labels as those used in our method.
\par
Table 2 shows the results in the non-overlapping setting. The mean Intersection over Union (IoU) over all the target-domain classes were used as the evaluation metric. For the incomplete versions of our method which trained multiple models but no final model, we independently evaluated the ensemble models composed of one backbone and multiple classifiers and reported the average performance. As shown in Table 2, the cross-model consistency and the adversarial learning, which are the two components of the model-invariant feature learning, successively improved the performance. By introducing the maximum squares loss, we obtained the results of Stage \Romannum{1} of our method with further improvements. Finally, we performed the model integration of Stage \Romannum{2} and achieved the best performance with significant improvements compared to the baseline trained with only the self-training. The above results for the ablation study validated the effectiveness of each component of our method.
\par
Compared to other adaptation methods, our method outperformed all the others even with only Stage \Romannum{1} in the non-overlapping setting. Moreover, both the SSMA and the MSMA methods need to infer with multiple models to obtain the complete predictions while the final model of our method has only one backbone and thus spends much less time for inference than the two methods. The MSMA method failed in the non-overlapping setting because the weighted model ensemble of the method is meaningless without any common classes.
\par
Figure 2 shows two examples of the qualitative results of our method and the source-domain models without adaptation using source domains of $S$+$G$. It can be seen in the figure that the segmentation of both background and foreground classes was improved by our method. In the upper example, the predictions for sky, sidewalk, and vegetation became clearly better after the adaptation. And in the lower example, the classes including sidewalk, traffic sign, vegetation, and rider were segmented more precisely in the result of our method.

\begin{table}[t]
\centering
\caption{Results in the partly-overlapping setting. ST: self-training. CMC: cross-model consistency. ADV: adversarial learning. MSL: maximum squares loss. MI: model integration. PM: proposed method.}
\setlength{\tabcolsep}{3.0pt}
\begin{tabular}{|l|ccc|}
\hline
Method & $S$+$G$ & $S$+$T$ & $G$+$T$ \\
\hline
ST & 44.2 & 44.0 & 39.1 \\
ST+CMC & 46.0 & 44.8 & 39.6 \\
ST+CMC+ADV & 46.6 & 45.2 & 40.6 \\
ST+CMC+ADV+MSL (\textbf{=Stage \Romannum{1} of PM}) & 47.4 & 45.9 & 42.2 \\
ST+CMC+ADV+MSL+MI (\textbf{=PM}) & \textbf{48.3} & \textbf{47.2} & \textbf{43.5} \\
\hline
SSMA~\cite{fleuret2021uncertainty} & 47.2 & 46.3 & 42.7 \\
MSMA~\cite{ahmed2021unsupervised} & 46.5 & 44.1 & 41.9 \\
UDA~\cite{tsai2018learning} & 47.9 & 46.9 & 41.6 \\
\hline
\end{tabular}
\end{table}

\begin{table}[t]
\centering
\caption{The per-class IoUs in the partly-overlapping setting for analyzing the influence of the model-invariant feature learning and the model integration on the performance over common and uncommon classes. ST: self-training. MIF: model-invariant feature learning, which is equal to ST+CMC+ADV in Table 3. S-\Romannum{1}: Stage \Romannum{1} of the proposed method. PM: proposed method.}
\tiny
\setlength{\tabcolsep}{1.2pt}
\begin{tabular}{|c|l|ccccccccccc|c|cccccccc|c|}
\hline
\multicolumn{2}{|c|}{} & \multicolumn{12}{c|}{IoUs of background classes (common)} & \multicolumn{9}{c|}{IoUs of foreground classes (uncommon)} \\
\cline{3-23}
\multicolumn{2}{|c|}{\makecell{Setting/\\Method}} & \rotatebox{90}{road} & \rotatebox{90}{side.} & \rotatebox{90}{buil.} & \rotatebox{90}{wall} & \rotatebox{90}{fence} & \rotatebox{90}{pole} & \rotatebox{90}{t-lig.} & \rotatebox{90}{t-sign} & \rotatebox{90}{vege.} & \rotatebox{90}{terr.} & \rotatebox{90}{sky} & mean& \rotatebox{90}{pers.} & \rotatebox{90}{rider} & \rotatebox{90}{car} & \rotatebox{90}{truck} & \rotatebox{90}{bus} & \rotatebox{90}{train} & \rotatebox{90}{moto.} & \rotatebox{90}{bicy.} & mean \\
\hline
\multirow{4}*{$S$+$G$} & ST & 81.1 & 43.7 & 79.4 & 28.0 & 19.7 & 38.0 & 42.3 & 45.6 & 84.2 & 32.1 & 80.3 & 52.2 & 58.3 & 25.5 & 80.0 & 18.9 & 16.3 & 1.7 & 20.3 & 45.2 & 33.3 \\
& MIF & 86.4 & 46.2 & 82.1 & 34.6 & 27.4 & 37.7 & 42.5 & 43.4 & 85.0 & 40.2 & 84.2 & 55.4 & 58.8 & 26.8 & 78.4 & 23.6 & 20.9 & 2.2 & 21.0 & 44.0 & 34.5 \\
& S-\Romannum{1} & 85.1 & 46.0 & 82.6 & 34.6 & 27.3 & 37.5 & 41.6 & 42.5 & 84.9 & 39.5 & 84.0 & 55.1 & 58.4 & 25.2 & 80.3 & 29.9 & 31.2 & 2.7 & 24.5 & 42.6 & 36.8 \\
& PM & 87.3 & 47.5 & 83.4 & 34.1 & 28.1 & 37.5 & 42.1 & 43.8 & 85.2 & 41.8 & 84.3 & 55.9 & 58.2 & 25.0 & 81.0 & 34.5 & 33.2 & 3.2 & 24.4 & 43.4 & 37.9 \\
\hline
\multirow{4}*{$S$+$T$} & ST & 73.8 & 36.2 & 82.1 & 26.3 & 19.4 & 36.8 & 37.2 & 38.7 & 84.4 & 27.6 & 82.7 & 49.6 & 60.8 & 22.3 & 82.1 & 22.2 & 18.3 & 1.9 & 27.5 & 55.9 & 36.4 \\
& MIF & 74.1 & 39.5 & 83.1 & 27.3 & 18.3 & 37.1 & 36.6 & 37.4 & 84.5 & 24.9 & 85.0 & 49.8 & 62.0 & 26.6 & 82.1 & 27.2 & 24.1 & 1.6 & 31.6 & 55.0 & 38.8 \\
& S-\Romannum{1} & 76.8 & 39.4 & 83.5 & 26.6 & 20.8 & 37.3 & 38.0 & 37.3 & 85.0 & 21.3 & 83.7 & 50.0 & 62.7 & 31.3 & 81.7 & 27.3 & 35.0  & 1.4 & 27.0 & 55.9 & 40.3 \\
& PM & 80.6 & 42.5 & 84.1 & 27.2 & 23.4 & 37.6 & 37.0 & 39.6 & 85.2 & 23.2 & 84.6 & 51.4 & 62.4 & 31.9 & 82.2 & 31.8 & 37.1 & 1.9 & 30.4 & 54.0 & 41.5 \\
\hline
\multirow{4}*{$G$+$T$} & ST & 67.9 & 29.1 & 83.4 & 28.1 & 23.5 & 32.6 & 29.0 & 32.2 & 79.2 & 14.2 & 84.3 & 45.8 & 31.7 & 19.5 & 72.9 & 33.2 & 12.8 & 0.0 & 21.8 & 47.5 & 29.9 \\
& MIF & 73.9 & 32.3 & 83.5 & 28.8 & 22.3 & 33.2 & 26.8 & 26.5 & 78.6 & 15.9 & 86.4 & 46.2 & 35.1 & 20.6 & 82.0 & 36.1 & 16.3 & 0.0 & 24.2 & 48.8 & 32.9 \\
& S-\Romannum{1} & 82.2 & 37.3 & 83.3 & 27.7 & 17.2 & 33.7 & 25.8 & 23.8 & 80.1 & 20.3 & 85.6 & 47.0 & 35.5 & 20.2 & 84.4 & 33.2 & 30.2 & 0.0 & 32.1 & 49.5 & 35.6 \\
& PM & 84.4 & 39.3 & 83.5 & 26.5 & 22.5 & 33.9 & 26.2 & 26.3 & 80.1 & 21.5 & 84.6 & 48.1 & 38.5 & 21.5 & 84.6 & 35.1 & 36.7 & 0.0 & 30.9 & 51.0 & 37.3 \\
\hline
\end{tabular}
\end{table}

\subsection{Results in Partly-overlapping Setting}
For the partly-overlapping setting, we conducted the same experiments as those of the non-overlapping setting except $S$+$G$+$T$. The results are shown in Table 3. Similar to the results of the ablation study in the non-overlapping setting, each component contributed to the improvement consistently. Since both common and uncommon classes of the source domains exist in the partly-overlapping setting, we also show the per-class IoUs in Table 4 to analyze the influence of the model-invariant feature learning and the model integration on the performance over common and uncommon classes. By comparing ``MIF'' with ``ST'' in Table 4, the effectiveness of the model-invariant feature learning for both common and uncommon classes was indicated by the clear improvements except for the common classes of $S$+$T$. We think the slight improvement for the common classes of $S$+$T$ was because $S$ has a much less domain gap with the target domain for the background classes than that of $T$, and consequently harmonizing the model characteristics derived from $S$ and $T$ is not helpful to the generalization in the target domain. Moreover, the comparison between ``S-\Romannum{1}'' and ``PM'' showed that the model integration of Stage \Romannum{2} also improved the performance over both common and uncommon classes.
\par
As to the comparisons to the other adaptation methods, our method achieved the best performance again as shown in Table 3. However, unlike the non-overlapping setting, the version with only Stage \Romannum{1} of our method failed to outperform the SSMA and the UDA methods. Due to the presence of the common classes, the SSMA method benefited considerably from the model ensemble for obtaining complete predictions, which, however, decreased the inference speed by several times. Similarly, with the presence of the common classes, the MSMA method achieved reasonable performance with the weighted model ensemble. The UDA method achieved close performance to ours in $S$+$G$ and $S$+$T$ but required the access to the source-domain data.

\begin{table}[t]
\centering
\caption{Results in the fully-overlapping setting. ST: self-training. CMC: cross-model consistency. ADV: adversarial learning. MSL: maximum squares loss. MI: model integration. PM: proposed method.}
\setlength{\tabcolsep}{3.0pt}
\begin{tabular}{|l|ccc|}
\hline
Method & $S$+$G$ & $S$+$T$ & $G$+$T$ \\
\hline
ST & 47.6 & 45.7 & 44.0 \\
ST+CMC & 49.4 & 46.6 & 44.7 \\
ST+CMC+ADV & 50.9 & 47.7 & 45.7 \\
ST+CMC+ADV+MSL (\textbf{=Stage \Romannum{1} of PM}) & 51.0 & 47.8 & 45.7 \\
ST+CMC+ADV+MSL+MI (\textbf{=PM}) & \textbf{51.7} & 48.7 & \textbf{46.2} \\
\hline
SSMA~\cite{fleuret2021uncertainty} & 49.8 & 47.6 & 45.5 \\
MSMA~\cite{ahmed2021unsupervised} & \textbf{51.7} & 47.2 & 44.2 \\
UDA~\cite{tsai2018learning} & 51.6 & \textbf{49.0} & 45.8 \\
\hline
\end{tabular}
\end{table}

\subsection{Results in Fully-overlapping Setting}
We conducted experiments in also the fully-overlapping setting which is exactly the general multi-source model adaptation. Due to the identical label space of the source domains and the target domain, we no longer performed model ensemble for the SSMA method and averaged the independent performances of all the trained models as the final performance. The MSMA method was evaluated with no changes since the weighted model ensemble acts as the core of the method. And for the UDA method, we used domain-specific discriminators but only one classifier. We did not conduct experiments in the $S$+$G$+$T$ setting since compared to the performance in $S$+$G$, no improvements can be gained by introducing $T$ due to the much larger domain gap between $T$ and the target domain.
\par
Table 5 shows the results in the fully-overlapping setting. Similarly to the other two settings, the results of the ablation study validated the effectiveness of each component of our method, with the only difference that the maximum squares loss made almost no contributions in the fully-overlapping setting. It can be explained by the fact that the pseudo labels generated in the fully-overlapping setting were more accurate than those in the other settings, which diminished the significance of the maximum squares loss. 
\par
In the comparisons to the other adaptation methods, our method still outperformed the SSMA method, while the UDA method achieved slightly better performance than ours in $S$+$T$. Moreover, the MSMA method achieved the same performance as ours in $S$+$G$ using the two source domains with closer domain gaps with the target domain, which indicated that the efficiency of the weighted model ensemble is maximized with source domains that have the same label space and similar domain gaps with the target domain. Overall, our method has the best cost-performance ratio considering the inference speed and the requirement for the access to source-domain data.

\section{Conclusion}
This paper has presented a novel problem named union-set multi-source model adaptation, which requires the union of the source-domain label spaces to be equal to the target-domain label space and is thus applicable to a wider range of practical scenarios than that with the general multi-source setting. To tackle the problem of union-set multi-source model adaptation for semantic segmentation, we proposed a method with a novel learning strategy, model-invariant feature learning, to improve the generalization in the target domain by harmonizing the diverse characteristics of the source-domain models. Moreover, we further performed the model integration which distills the knowledge from the adapted models and trains a final model with improved performance. The effectiveness of each component of our method was validated by the results of the elaborate ablation studies, and the superiority of our method compared to previous adaptation methods was demonstrated by the results of the extensive experiments in various settings.

\clearpage
%
%
\bibliographystyle{splncs04}
\bibliography{refs}

\begin{thebibliography}{10}
\providecommand{\url}[1]{\texttt{#1}}
\providecommand{\urlprefix}{URL }
\providecommand{\doi}[1]{https://doi.org/#1}

\bibitem{ahmed2021unsupervised}
Ahmed, S.M., Raychaudhuri, D.S., Paul, S., Oymak, S., Roy-Chowdhury, A.K.:
  Unsupervised multi-source domain adaptation without access to source data.
  In: Proceedings of the IEEE/CVF Conference on Computer Vision and Pattern
  Recognition. pp. 10103--10112 (2021)

\bibitem{chen2017deeplab}
Chen, L.C., Papandreou, G., Kokkinos, I., Murphy, K., Yuille, A.L.: Deeplab:
  Semantic image segmentation with deep convolutional nets, atrous convolution,
  and fully connected crfs. IEEE Transactions on Pattern Analysis and Machine
  Intelligence  \textbf{40}(4),  834--848 (2017)

\bibitem{chen2019domain}
Chen, M., Xue, H., Cai, D.: Domain adaptation for semantic segmentation with
  maximum squares loss. In: Proceedings of the IEEE/CVF International
  Conference on Computer Vision. pp. 2090--2099 (2019)

\bibitem{chen2019crdoco}
Chen, Y.C., Lin, Y.Y., Yang, M.H., Huang, J.B.: Crdoco: Pixel-level domain
  transfer with cross-domain consistency. In: Proceedings of the IEEE/CVF
  conference on computer vision and pattern recognition. pp. 1791--1800 (2019)

\bibitem{choi2019self}
Choi, J., Kim, T., Kim, C.: Self-ensembling with gan-based data augmentation
  for domain adaptation in semantic segmentation. In: Proceedings of the
  IEEE/CVF International Conference on Computer Vision. pp. 6830--6840 (2019)

\bibitem{cordts2016cityscapes}
Cordts, M., Omran, M., Ramos, S., Rehfeld, T., Enzweiler, M., Benenson, R.,
  Franke, U., Roth, S., Schiele, B.: The cityscapes dataset for semantic urban
  scene understanding. In: Proceedings of the IEEE Conference on Computer
  Vision and Pattern Recognition. pp. 3213--3223 (2016)

\bibitem{du2019ssf}
Du, L., Tan, J., Yang, H., Feng, J., Xue, X., Zheng, Q., Ye, X., Zhang, X.:
  Ssf-dan: Separated semantic feature based domain adaptation network for
  semantic segmentation. In: Proceedings of the IEEE/CVF International
  Conference on Computer Vision. pp. 982--991 (2019)

\bibitem{fleuret2021uncertainty}
Fleuret, F., et~al.: Uncertainty reduction for model adaptation in semantic
  segmentation. In: Proceedings of the IEEE/CVF Conference on Computer Vision
  and Pattern Recognition. pp. 9613--9623 (2021)

\bibitem{ganin2015unsupervised}
Ganin, Y., Lempitsky, V.: Unsupervised domain adaptation by backpropagation.
  In: Proceedings of the International Conference on Machine Learning. pp.
  1180--1189 (2015)

\bibitem{goodfellow2014generative}
Goodfellow, I., Pouget-Abadie, J., Mirza, M., Xu, B., Warde-Farley, D., Ozair,
  S., Courville, A., Bengio, Y.: Generative adversarial nets. Advances in
  Neural Information Processing Systems  \textbf{27} (2014)

\bibitem{he2021multi}
He, J., Jia, X., Chen, S., Liu, J.: Multi-source domain adaptation with
  collaborative learning for semantic segmentation. In: Proceedings of the
  IEEE/CVF Conference on Computer Vision and Pattern Recognition. pp.
  11008--11017 (2021)

\bibitem{he2016deep}
He, K., Zhang, X., Ren, S., Sun, J.: Deep residual learning for image
  recognition. In: Proceedings of the IEEE Conference on Computer Vision and
  Pattern Recognition. pp. 770--778 (2016)

\bibitem{hoffman2018algorithms}
Hoffman, J., Mohri, M., Zhang, N.: Algorithms and theory for multiple-source
  adaptation. Advances in Neural Information Processing Systems  \textbf{31}
  (2018)

\bibitem{hoffman2018cycada}
Hoffman, J., Tzeng, E., Park, T., Zhu, J.Y., Isola, P., Saenko, K., Efros, A.,
  Darrell, T.: Cycada: Cycle-consistent adversarial domain adaptation. In:
  Proceedings of the International Conference on Machine Learning. pp.
  1989--1998 (2018)

\bibitem{li2020model}
Li, R., Jiao, Q., Cao, W., Wong, H.S., Wu, S.: Model adaptation: Unsupervised
  domain adaptation without source data. In: Proceedings of the IEEE/CVF
  Conference on Computer Vision and Pattern Recognition. pp. 9641--9650 (2020)

\bibitem{li2019bidirectional}
Li, Y., Yuan, L., Vasconcelos, N.: Bidirectional learning for domain adaptation
  of semantic segmentation. In: Proceedings of the IEEE/CVF Conference on
  Computer Vision and Pattern Recognition. pp. 6936--6945 (2019)

\bibitem{liang2020we}
Liang, J., Hu, D., Feng, J.: Do we really need to access the source data?
  source hypothesis transfer for unsupervised domain adaptation. In:
  Proceedings of the International Conference on Machine Learning. pp.
  6028--6039 (2020)

\bibitem{liu2021source}
Liu, Y., Zhang, W., Wang, J.: Source-free domain adaptation for semantic
  segmentation. In: Proceedings of the IEEE/CVF Conference on Computer Vision
  and Pattern Recognition. pp. 1215--1224 (2021)

\bibitem{long2015learning}
Long, M., Cao, Y., Wang, J., Jordan, M.: Learning transferable features with
  deep adaptation networks. In: Proceedings of the International Conference on
  Machine Learning. pp. 97--105 (2015)

\bibitem{luo2019taking}
Luo, Y., Zheng, L., Guan, T., Yu, J., Yang, Y.: Taking a closer look at domain
  shift: Category-level adversaries for semantics consistent domain adaptation.
  In: Proceedings of the IEEE/CVF Conference on Computer Vision and Pattern
  Recognition. pp. 2507--2516 (2019)

\bibitem{peng2019moment}
Peng, X., Bai, Q., Xia, X., Huang, Z., Saenko, K., Wang, B.: Moment matching
  for multi-source domain adaptation. In: Proceedings of the IEEE/CVF
  International Conference on Computer Vision. pp. 1406--1415 (2019)

\bibitem{richter2016playing}
Richter, S.R., Vineet, V., Roth, S., Koltun, V.: Playing for data: Ground truth
  from computer games. In: Proceedings of the European Conference on Computer
  Vision. pp. 102--118 (2016)

\bibitem{ros2016synthia}
Ros, G., Sellart, L., Materzynska, J., Vazquez, D., Lopez, A.M.: The synthia
  dataset: A large collection of synthetic images for semantic segmentation of
  urban scenes. In: Proceedings of the IEEE Conference on Computer Vision and
  Pattern Recognition. pp. 3234--3243 (2016)

\bibitem{stan2021unsupervised}
Stan, S., Rostami, M.: Unsupervised model adaptation for continual semantic
  segmentation. In: Proceedings of the AAAI Conference on Artificial
  Intelligence. vol.~35, pp. 2593--2601 (2021)

\bibitem{tsai2018learning}
Tsai, Y.H., Hung, W.C., Schulter, S., Sohn, K., Yang, M.H., Chandraker, M.:
  Learning to adapt structured output space for semantic segmentation. In:
  Proceedings of the IEEE Conference on Computer Vision and Pattern
  Recognition. pp. 7472--7481 (2018)

\bibitem{tsai2019domain}
Tsai, Y.H., Sohn, K., Schulter, S., Chandraker, M.: Domain adaptation for
  structured output via discriminative patch representations. In: Proceedings
  of the IEEE/CVF International Conference on Computer Vision. pp. 1456--1465
  (2019)

\bibitem{vu2019advent}
Vu, T.H., Jain, H., Bucher, M., Cord, M., P{\'e}rez, P.: Advent: Adversarial
  entropy minimization for domain adaptation in semantic segmentation. In:
  Proceedings of the IEEE/CVF Conference on Computer Vision and Pattern
  Recognition. pp. 2517--2526 (2019)

\bibitem{wrenninge2018synscapes}
Wrenninge, M., Unger, J.: Synscapes: A photorealistic synthetic dataset for
  street scene parsing. arXiv preprint arXiv:1810.08705  (2018)

\bibitem{wu2018dcan}
Wu, Z., Han, X., Lin, Y.L., Uzunbas, M.G., Goldstein, T., Lim, S.N., Davis,
  L.S.: Dcan: Dual channel-wise alignment networks for unsupervised scene
  adaptation. In: Proceedings of the European Conference on Computer Vision.
  pp. 518--534 (2018)

\bibitem{xia2021adaptive}
Xia, H., Zhao, H., Ding, Z.: Adaptive adversarial network for source-free
  domain adaptation. In: Proceedings of the IEEE/CVF International Conference
  on Computer Vision. pp. 9010--9019 (2021)

\bibitem{yang2020unsupervised}
Yang, S., Wang, Y., van~de Weijer, J., Herranz, L., Jui, S.: Unsupervised
  domain adaptation without source data by casting a bait. arXiv preprint
  arXiv:2010.12427  (2020)

\bibitem{yang2020fda}
Yang, Y., Soatto, S.: Fda: Fourier domain adaptation for semantic segmentation.
  In: Proceedings of the IEEE/CVF Conference on Computer Vision and Pattern
  Recognition. pp. 4085--4095 (2020)

\bibitem{yao2021multi}
Yao, X., Zhao, S., Xu, P., Yang, J.: Multi-source domain adaptation for object
  detection. In: Proceedings of the IEEE/CVF International Conference on
  Computer Vision. pp. 3273--3282 (2021)

\bibitem{zhang2018fully}
Zhang, Y., Qiu, Z., Yao, T., Liu, D., Mei, T.: Fully convolutional adaptation
  networks for semantic segmentation. In: Proceedings of the IEEE Conference on
  Computer Vision and Pattern Recognition. pp. 6810--6818 (2018)

\bibitem{zhao2018adversarial}
Zhao, H., Zhang, S., Wu, G., Moura, J.M., Costeira, J.P., Gordon, G.J.:
  Adversarial multiple source domain adaptation. Advances in Neural Information
  Processing Systems  \textbf{31},  8559--8570 (2018)

\bibitem{zhao2019multi}
Zhao, S., Li, B., Yue, X., Gu, Y., Xu, P., Hu, R., Chai, H., Keutzer, K.:
  Multi-source domain adaptation for semantic segmentation. Advances in Neural
  Information Processing Systems  \textbf{32},  7287--7300 (2019)

\bibitem{zheng2021rectifying}
Zheng, Z., Yang, Y.: Rectifying pseudo label learning via uncertainty
  estimation for domain adaptive semantic segmentation. International Journal
  of Computer Vision  \textbf{129}(4),  1106--1120 (2021)

\bibitem{zhu2017unpaired}
Zhu, J.Y., Park, T., Isola, P., Efros, A.A.: Unpaired image-to-image
  translation using cycle-consistent adversarial networks. In: Proceedings of
  the IEEE International Conference on Computer Vision. pp. 2223--2232 (2017)

\bibitem{zou2018unsupervised}
Zou, Y., Yu, Z., Kumar, B., Wang, J.: Unsupervised domain adaptation for
  semantic segmentation via class-balanced self-training. In: Proceedings of
  the European Conference on Computer Vision. pp. 289--305 (2018)

\end{thebibliography}
\end{document}